\documentclass[sigconf]{acmart}

\usepackage{arydshln}
\usepackage{amsmath,amsfonts}
\usepackage{algorithmic}
\usepackage{algorithm}
\usepackage{array}
\usepackage[caption=false]{subfig}
\usepackage{textcomp}
\usepackage{stfloats}
\usepackage{url}
\usepackage{verbatim}
\usepackage{graphicx}
\usepackage{booktabs}
\usepackage{multirow}
\usepackage{enumitem}
\usepackage{soul}
\usepackage{caption}
\AtBeginDocument{%
  }

\copyrightyear{2024}
\acmYear{2024}
\setcopyright{acmlicensed}\acmConference[CIKM '24]{Proceedings of the 33rd ACM International Conference on Information and Knowledge Management}{October 21--25, 2024}{Boise, ID, USA}
\acmBooktitle{Proceedings of the 33rd ACM International Conference on Information and Knowledge Management (CIKM '24), October 21--25, 2024, Boise, ID, USA}
\acmDOI{10.1145/3627673.3680261}
\acmISBN{979-8-4007-0436-9/24/10}

\begin{document}

\title{Towards Effective Fusion and Forecasting of Multimodal Spatio-temporal Data for Smart Mobility}

\author{Chenxing Wang}
\authornote{Supervised by Fang Zhao and Haiyong Luo (zfsse@bupt.edu.cn; yhluo@ict.ac.cn).}
\email{wangchenxing@bupt.edu.cn}
\orcid{0000-0003-4096-7972}
\affiliation{%
  \institution{School of Computer Science, Beijing University of Posts and Telecommunications}
  \city{Beijing}
  \country{China}
}

\renewcommand{\shortauthors}{Chenxing Wang}

\begin{abstract}
With the rapid development of location based services, multimodal spatio-temporal (ST) data including trajectories, transportation modes, traffic flow and social check-ins are being collected for deep learning based methods. These deep learning based methods learn ST correlations to support the downstream tasks in the fields such as smart mobility, smart city and other intelligent transportation systems. Despite their effectiveness, ST data fusion and forecasting methods face practical challenges in real-world scenarios. 
First, forecasting performance for ST data-insufficient area is inferior, making it necessary to transfer meta knowledge from heterogeneous area to enhance the sparse representations.
Second, it is nontrivial to accurately forecast in multi-transportation-mode scenarios due to the fine-grained ST features of similar transportation modes, making it necessary to distinguish and measure the ST correlations to alleviate the influence caused by entangled ST features.
At last, partial data modalities (e.g., transportation mode) are lost due to privacy or technical issues in certain scenarios, making it necessary to effectively fuse the multimodal sparse ST features and enrich the ST representations. To tackle these challenges, our research work aim to develop effective fusion and forecasting methods for multimodal ST data in smart mobility scenario. In this paper, we will introduce our recent works that investigates the challenges in terms of various real-world applications and establish the open challenges in this field for future work.
\end{abstract}



\maketitle

\section{Introduction}

With the acceleration of urbanization, multimodalspatio-temporal (ST) data fusion and forecasting techniques for smart mobility scenarios are increasingly becoming a research hotspot. Currently, previous studies \cite{DBLP:conf/icde/FangQL0XZ023,DBLP:conf/sigir/QinFL0022,DBLP:journals/tkde/FangQLZZ24,DBLP:conf/icde/0018Z0Z23,DBLP:conf/kdd/ChenXGFMCC22} have shown high prediction and recognition performance in homogeneous urban unimodal scenes. However, when dealing with heterogeneous, fine-grained and multimodal sparse scenes, ST data fusion and forecasting face many challenges, such as the heterogeneous of ST features in different regions, the similarity entanglement of fine-grained ST features, and the sparsity of partial modalities data in multimodal scenes, which all significantly degrade the performance of ST forecasting. To this end, our research aims to explore and solve the above challenges to improve the performance of ST data fusion and forecasting in smart mobility scenarios, especially in heterogeneous, fine-grained, and multimodal sparse scenarios.

\section{Preliminary}
\label{sec:pre}

To better understand our proposed research problem, we first introduce the research scope with illustrations and then introduce the problem definitions of scenarios related to our research.

\subsection{Research scope, questions and scenarios}

\begin{figure*}[htbp] \centering \includegraphics[width=0.7\linewidth, trim= 0cm 1cm 0cm 0cm, clip]{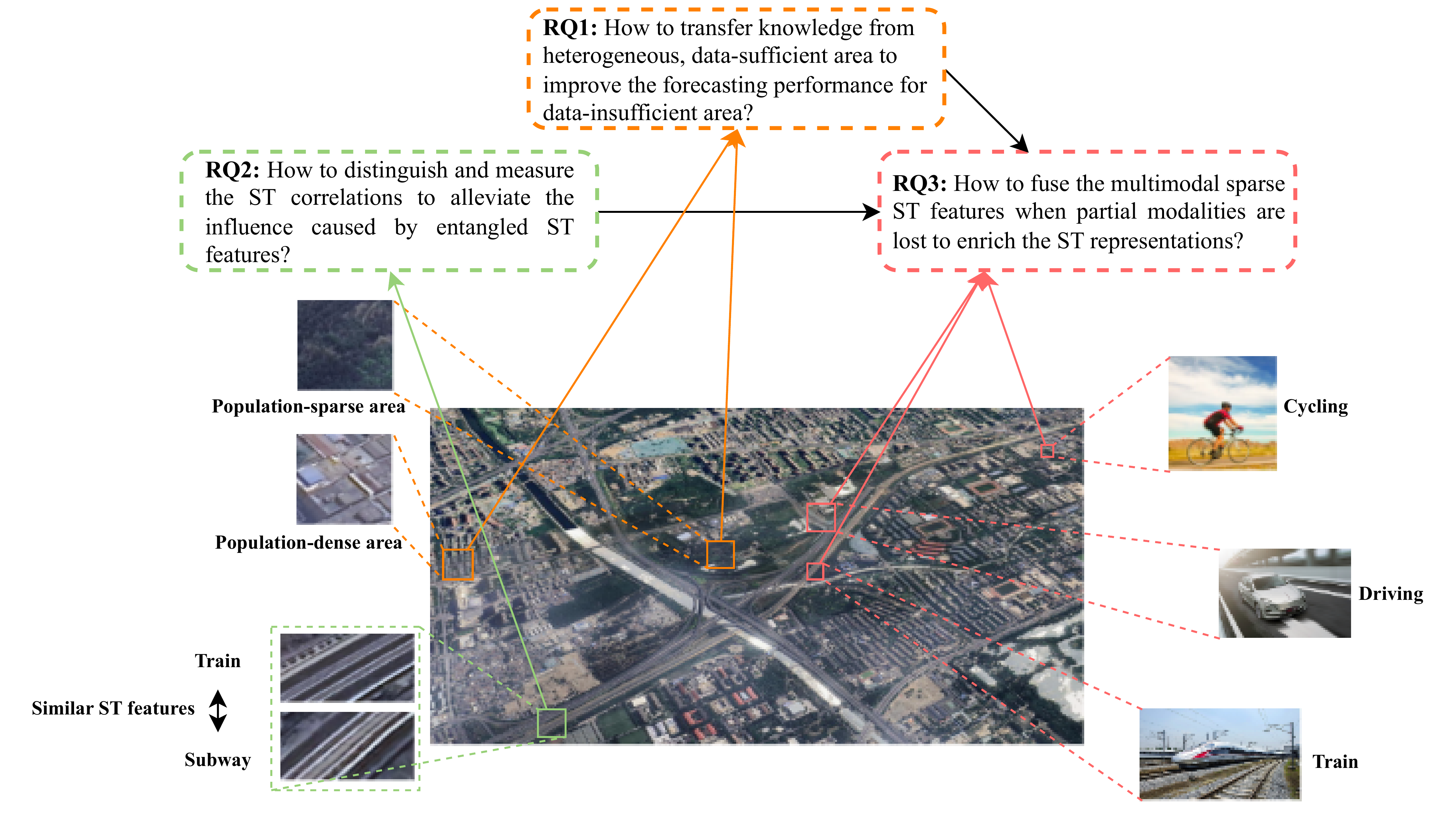}
\caption{Our research scope and related scenarios. }\label{fig:scope}  
\end{figure*}

As illustrated in Figure \ref{fig:scope}, our research scope mainly contains three research questions (RQ), which deals with \textit{heterogeneous}, \textit{fine-grained} and \textit{multimodal sparse} scenes, respectively. 
\begin{itemize}[leftmargin=*,topsep=0pt,noitemsep]
    \item \textbf{RQ1:} Existing ST forecasting methods perform well with the support of large-scale spatio-temporal data, but the performance degradation is obvious in cities and regions with less population or ST data \cite{DBLP:conf/dasfaa/FanXZL22}. The critical to solve this problem is how to effectively extract potential ST knowledge of cities, learn heterogeneous ST correlations among different cities, and gradually transfer this rich meta-knowledge to urban regions with less ST data to improve the performance of ST forecasting methods in real applications.
    \item \textbf{RQ2:} Accurate ST forecasting in multi-transportation-mode scenario is a challenging task because fine-grained transportation modes have similar and entangled ST characteristics that are difficult to accurately distinguish \cite{DBLP:journals/tits/0001L0Q21}. The key point to answer this research question is how to effectively measure the complex ST correlations and reduce the impact of fine-grained ST feature entanglement to enhance the forecasting performance in multi-transportation-mode scenario.
    \item \textbf{RQ3:} Existing ST data forecasting algorithms usually rely on quantities of annotated data, but in urban scenarios, partial modalities (e.g., transportation mode) tend to be sparse due to privacy and technical limitations \cite{DBLP:conf/dasfaa/XuXZLLL20}. Our research focus on how to effectively fuse multimodal sparse ST features in urban scenes and enrich ST feature representations in order to improve the model's performance in real-world scenario.
\end{itemize}

To answer these research questions, we mainly focus on three scenarios, including \textit{multi-region travel time estimation}, \textit{multi-transportation-mode travel time estimation} and \textit{multi-transportation-mode trajectory recovery}. We then introduce their problem formulations, respectively.

\begin{figure}
\centering 
\subfloat[Route TTE.]{\includegraphics[width=0.7\linewidth, trim= 0cm 0cm 0cm 0cm, clip]{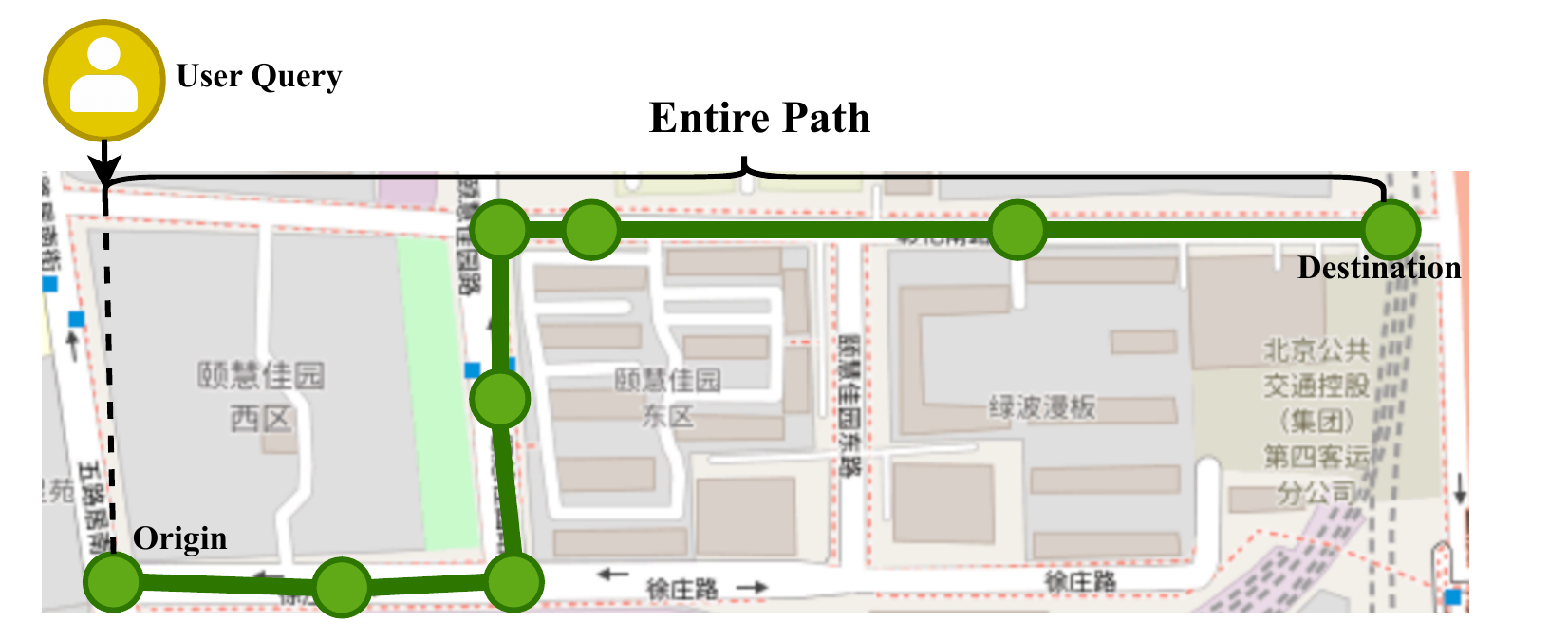}
\label{fig:example_route_tte}} 
\vfil 
\subfloat[En Route TTE.]{\includegraphics[width=0.7\linewidth, trim= 0cm 0cm 0cm 0cm, clip]{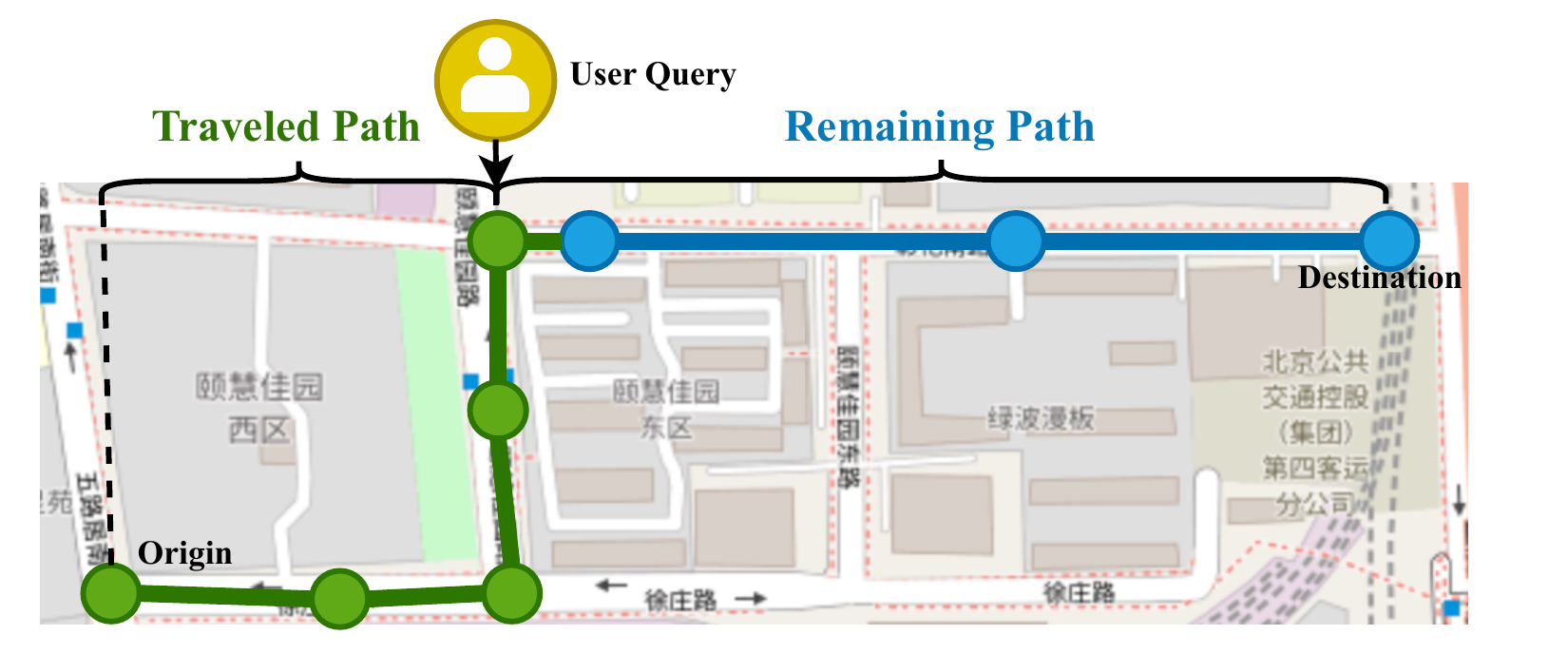}
\label{fig:example_en_route_tte} } 
\caption{Illustration of Route TTE and En Route TTE.}~\label{fig:example_route_tte_vs_en_routette}
\end{figure}

\subsection{Multi-region Route Travel Time Estimation (MRRTTE)}

As illustrated in Figure \ref{fig:example_route_tte}, when user sends the query at departure time $t_d$ for a given route $r$, the oracle will output the estimation through the deep learning model. It has been widely investigated for single-region route travel time estimation \cite{DBLP:journals/pacmmod/LinWHGYLJ23,DBLP:conf/kdd/LiuJLC23,DBLP:journals/tits/ZouLMTFL23,DBLP:conf/kdd/ChenXGFMCC22} recently. Nevertheless, multi-region route travel time estimation has not been well explored in the literature. Since the knowledge transfer of ST features is reasonable \cite{DBLP:conf/www/YaoLWTL19}, we hope to propose the meta learning based model to learn critical knowledge for estimation based on data-sufficient region $\mathcal{R}_s$ and then enhance the ST representations for those data-insufficient region $\mathcal{R}_{is}$ so that its performance can be improved.

\subsection{Multi-transportation-mode En Route Travel Time Estimation (MTERTTE)}

As illustrated in Figure \ref{fig:example_en_route_tte}, different from route travel time estimation, when user sends the query at departure time $t_d$, not only the route $r$ for the remaining path but also the traveled path $r_t$ are sent to the oracle to output the estimation. Existing works employ meta learning to transfer knowledge from the traveled the path to the remaining path in few shot learning scheme which limits the efficiency \cite{DBLP:conf/ijcai/Fan00000L22,DBLP:conf/kdd/FangHWLSW21}. Moreover, these works only consider the vehicle-based (i.e., cars) queries which cannot be applied to queries that consist of multiple transportation modes. It is nontrivial to achieve this goal, since the entangled ST features between similar transportation modes are difficult to distinguish and measure. For example, train and subway modes (or car and bus modes) are difficult to distinguish in relatively short time window \cite{DBLP:journals/tits/0001L0Q21}.
To this end, in our research, we hope to propose a solution that estimates the multi-transportation-mode en route travel time for given queries that can disentangle the ST features for better performance.

\subsection{Multi-transportation-mode Trajectory Recovery (MTRec)}

 \begin{figure}[htbp] \centering \includegraphics[width=7cm]{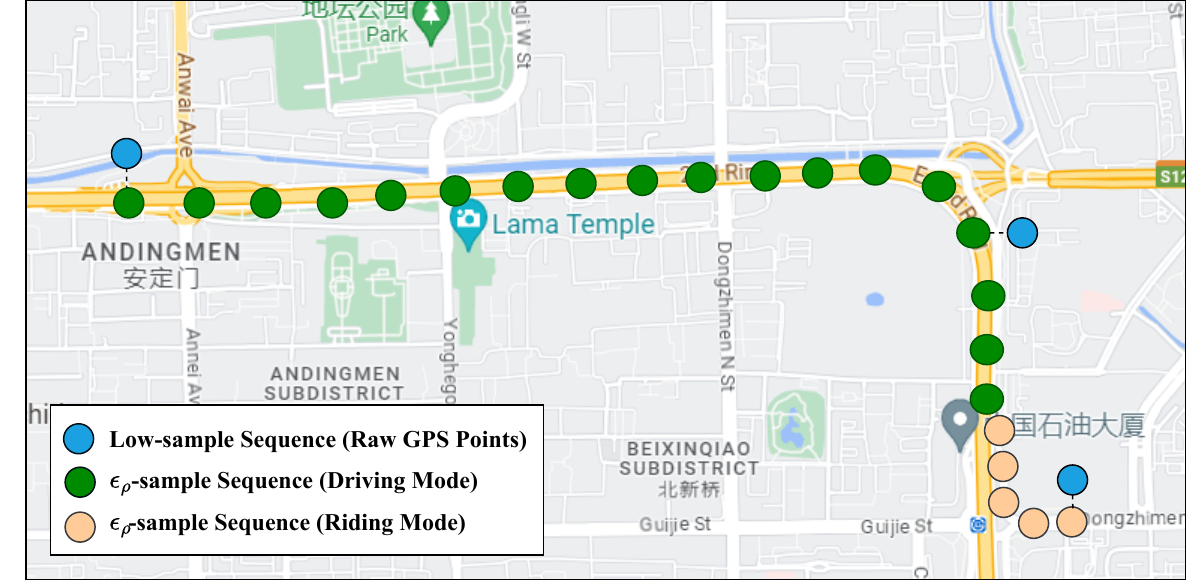}
\caption{Illustration of multi-transportation-mode trajectory recovery.}\label{fig:ptrajrec} \end{figure}

As illustrated in Figure \ref{fig:ptrajrec}, given the low-sample sequence in multiple transportation modes, the oracle will output the recovered $\epsilon_p$-sample sequence. Previous works \cite{DBLP:conf/kdd/RenRL0ML021,DBLP:conf/icde/0018Z0Z23} treat this task as the vehicle-only recovery which ignores the inherent heterogeneity. Moreover, some modalities (e.g., the transportation modes) tend to be sparse due to privacy or technical issues which brings difficulties to fusion of multimodal ST features. To this end, we hope to propose a deep learning based solution to tackle the fusion issues caused by sparsity so that multimodal ST features can be enriched and the recovery performance can be improved.

\section{Proposed methods and results}


\subsection{MRRTTE for RQ1}

Most existing MRRTTE rely on large amounts of historical trajectory data to construct deep learning models to predict travel time in specific urban areas. These algorithms often rely on spatio-temporally dense datasets for training to achieve a high degree of generalization in a specific urban area. However, the historical trajectory data that can be collected in less densely populated areas are usually insufficient to train predictive models with high robustness. To solve this problem, our research utilizes meta-learning methods to comprehensively analyze historical trajectory data from different urban areas and design a meta-learning-based deep learning algorithm called MetaTTE to learn spatio-temporal features in urban areas and migrating this meta-knowledge to data-sparse urban areas in order to improve the accuracy of the estimation. Our proposed the MetaTTE achieves higher accuracy compared with the baselines (see Table \ref{tab:mrrtte}) and the detailed results can be found at \cite{DBLP:journals/tits/WangZZLQF22}.

\begin{table} \caption{Performance comparison for MRRTTE.} \centering
\resizebox{1.0\linewidth}{!}{\begin{tabular}{l|c|c|c|c|c|c} \toprule 
 \multirow{2}{*}{Baselines} & \multicolumn{3}{c}{Chengdu} &
\multicolumn{3}{c}{Porto} \\ & MAE & MAPE (\%) & RMSE & MAE & MAPE (\%) & RMSE\\ 
\midrule 
AVG & 442.20 & 39.71 & 8443.60 & 182.64 & 26.66 & 1128.21 \\ 
LR & 516.23 & 49.09 & 1204.99 & 194.40 & 33.90 & 279.20\\
GBM & 454.50 & 41.67 & 1121.32 & \underline{148.53} & \underline{24.59} & \underline{209.07}\\
TEMP & \underline{334.60} & \underline{39.70} & \underline{761.05} & 174.44 & 28.73 & 260.81\\ 
\midrule 
WDR & 433.99 & 29.74 & 1024.92 & 164.04 & 22.84 & 244.41\\
DeepTTE & 413.09 & \underline{24.22} & \underline{926.04} & \underline{84.29} & \underline{14.79} & \underline{\textbf{90.29}}\\
STNN & 427.33 & 30.08 & 1011.88 & 226.30 & 35.44 & 331.75\\ 
MURAT & \underline{396.01} & 29.29 & 994.95  & 165.91 & 27.10 & 177.83\\ 
Nei-TTE  & 414.16 & 30.04 & 1038.71 & 106.30 & 15.23 & 183.03\\
\midrule 
MetaTTE (ours) & \underline{\textbf{236.38}} & \underline{\textbf{23.69}} & \textbf{\underline{745.11}} & \underline{\textbf{62.43}} & \underline{\textbf{8.83}} & \underline{196.78}\\ \bottomrule
\end{tabular}}
\label{tab:mrrtte} \end{table}

\subsection{MTERTTE for RQ2}

MTERTTE faces multiple challenges, including entangled spatio-temporal features of fine-grained transportation modes as well as diverse user behaviors and travel intentions, which make it difficult to achieve accurate estimation. Particularly, in practical applications, certain transportation modes are prone to cause confusion in model differentiation due to the possession of entangled spatio-temporal features, resulting in insufficient distance between classes to differentiate them. To cope with the above problems, we design an attention based hybrid transportation mode modeling approach. Specifically, we design an algorithm based on autocorrelation attention mechanism for predicting the travel time of multiple transportation modes. The sub-sequence correlations in multimodal spatio-temporal data are learnt through the autocorrelation attention mechanism, and we carry out personalized embedding and feature learning for the user's personalized mode characteristics, and then conduct the adaptive fusion of multimodal spatio-temporal features, and finally, we design a multi-tasking framework to introduce spatio-temporal features of the transportation modes to assist in supervising the training optimization which helps achieve the accurate estimation.

\subsection{MTRec for RQ3}

In our research, we focus on the transportation-aware trajectory recovery problem, which is distinct from the conventional vehicle-based trajectory recovery, facing two major challenges: heterogeneity and personalization. For the heterogeneity, the velocity of the mobile object is intrinsically correlated with the specific transportation mode, containing inherent heterogeneity. For the personalization, the trajectory data is complicated by substantial variations in users, which are different in personalized behaviors. To address these challenges, we design a novel effective multi-modal deep model, coined as PTrajRec, for transportation-aware trajectory recovery. Specifically, we initially embed location, behavior, and transportation mode modalities in distinct channels, which not only reflect spatio-temporal information encapsulated in location sequences but also introduce the heterogeneity and personalization characteristics associated with mode and behavior sequences. For further modeling these modalities, we employ the auto-correlation mechanism to learn periodic dependencies on the temporal dimension and the graph attention mechanism to learn road network dependencies on the spatial dimension. At last, we propose a dual-view constraint mechanism to assist the fine-grained trajectory recovery and design three auxiliary tasks to address the inherent heterogeneity. The experimental results in Table \ref{tab:ptrajrec} demonstrates its superior.

\begin{table*}[htbp] \caption{Performance comparison of MTRec.} 
\aboverulesep=0ex
\belowrulesep=0ex
\centering
\resizebox{0.7\linewidth}{!}{\begin{tabular}{l|ccccc|ccccc} 
\toprule 
 \multirow{2}{*}{Baselines} & \multicolumn{5}{c|}{Geolife} &
\multicolumn{5}{c}{Singapore} \\
\cmidrule(lr){2-6} \cmidrule(lr){7-11}
& Recall $\uparrow$ & Precision $\uparrow$  & Accuracy $\uparrow$ & MAE $\downarrow$ & RMSE $\downarrow$ & Recall $\uparrow$ & Precision $\uparrow$ & Accuracy $\uparrow$ & MAE $\downarrow$ & RMSE $\downarrow$\\ 
\midrule
Linear & 0.2519 & 0.3822 & 0.2139 & 0.902 & 1.292 & 0.6456 & 0.6260 & 0.5635 & 0.967 & 1.536\\
DHTR & 0.2654 & 0.4301 & 0.2149 & 0.713 & 1.019 & 0.6556 & 0.7520 & 0.5668 & 0.874 & 0.964\\
\midrule
T3S & 0.2877 & 0.4883 & 0.2592 & 0.471 & 0.786 & 0.6884 & 0.7815 & 0.5816 & 0.600 & 0.944 \\
T2Vec & 0.2720 & 0.4878 & 0.2437 & 0.489 & 0.668 & 0.6816 & 0.7807 & 0.5810 & 0.642 & 0.983 \\
Transformer & 0.2641 & 0.4772 & 0.2408 & 0.534 & 0.722 & 0.6695 & 0.7691 & 0.5711 & 0.620 & 0.993 \\
NeuTraj & 0.3206 & 0.4125 & 0.2448 & 0.488 & 0.668 & 0.6748 & 0.7704 & 0.5548 & 0.628 & 1.001 \\
GTS & 0.3166 & 0.4697 & 0.2517 & 0.492 & 0.667 & 0.6664 & 0.7596 & 0.5705 & 0.633 & 1.000 \\
MTrajRec & 0.3267 & 0.4764 & 0.2616 & 0.440 & 0.601 & 0.6829 & 0.7928 & 0.5955 & 0.540 & 0.881 \\
RNTrajRec & \underline{0.3300} & \underline{0.4911} &  \underline{0.2630} & \underline{0.414} & \underline{0.575} & \underline{0.6946} & \underline{0.7942} &  \underline{0.5961} & \underline{0.533} & \underline{0.862} \\
\midrule
PTrajRec (ours) & \textbf{0.4351} & \textbf{0.5418} & \textbf{0.3824} & \textbf{0.266} & \textbf{0.358} &\textbf{0.7278} & \textbf{0.8544} & \textbf{0.6746} & \textbf{0.487} & \textbf{0.744} \\
\midrule
\bottomrule
\end{tabular}}
\label{tab:ptrajrec} \end{table*}

\section{Conclusion and future direction}

In general, the topic for heterogeneous, fine-grained and multimodal sparse characteristics of ST data are still open challenges for future works. In the era of foundation models, we would like to share, at least, the following research directions:
\begin{itemize}[leftmargin=*,topsep=0pt,noitemsep]
\item Multimodal foundation models for ST data that address the heterogeneity.
\item Finetuning and knowledge transfer based on LLM to support ST downstream tasks.
\item Effective prompt learning for ST downstream tasks to enrich the original ST feature representations.
\end{itemize}

\section*{Biography}

\textbf{Chenxing Wang} is a Ph.D student at School of Computer Science, Beijing University of Posts and Telecommunications. His research interests mainly focus on spatio-temporal data mining including but not limited to: trajectory recovery, travel time estimation, transportation mode detection, traffic flow forecasting and next POI recommendation. He has published several papers in top journals and conference proceedings, such as ICDE, SIGIR, and T-ITS. Additionally, he is now serving as the research intern at Weixin Group, Tencent supported by 2024 Tencent Rhino-bird Research Elite Program.

\begin{acks}
This work was supported in part by the National Natural Science Foundation of China under Grant 62261042, the Key Research Projects of the Joint Research Fund for Beijing Natural Science Foundation and the Fengtai Rail Transit Frontier Research Joint Fund under Grant L221003, Beijing Natural Science Foundation under Grant 4232035 and 4222034, the Strategic Priority Research Program of Chinese Academy of Sciences under Grant XDA28040500 and BUPT Excellent Ph.D. Students Foundation under Grant CX2022132.
\end{acks}

\bibliographystyle{ACM-Reference-Format}
\bibliography{main}

\end{document}